\pdfoutput=1
\documentclass[10pt,twocolumn,letterpaper]{article}

\usepackage{iccv}
\usepackage{times}
\usepackage{epsfig}
\usepackage{graphicx}
\usepackage{amsmath}
\usepackage{amssymb}

\usepackage{booktabs}
\usepackage{multirow}

\usepackage{tabularx}

\newcolumntype{Y}{>{\centering\arraybackslash}X}

\newcolumntype{f}[1]{>{\centering\arraybackslash\hspace{0pt}}p{#1}}

\usepackage{subcaption}
\let\oldforall\forall
\renewcommand{\forall}{\oldforall \; }

\usepackage{mathtools}
\DeclarePairedDelimiter{\ceil}{\lceil}{\rceil}
\DeclarePairedDelimiter{\floor}{\lfloor}{\rfloor}

\def \paravspace {-0\baselineskip}

\usepackage[pagebackref=true,breaklinks=true,colorlinks,bookmarks=false]{hyperref}

\iccvfinalcopy %

\def\iccvPaperID{3807} %

\ificcvfinal\fi

\begin{document}

\title{Unified Graph Structured Models for Video Understanding}

\author{Anurag Arnab \quad Chen Sun \quad Cordelia Schmid \\
Google Research \\
{\tt\small \{aarnab, chensun, cordelias\}@google.com}
}

\maketitle
\ificcvfinal\thispagestyle{empty}\fi

\begin{abstract}

Accurate video understanding involves reasoning about the relationships between actors, objects and their environment, often over long temporal intervals.
In this paper, we propose a message passing graph neural network that explicitly models these spatio-temporal relations and can use explicit representations of objects, when supervision is available, and implicit representations otherwise.
Our formulation generalises previous structured models for video understanding, 
and allows us to study how different design choices in graph structure and representation affect the model's performance.
We demonstrate our method on two different tasks requiring relational reasoning in videos -- spatio-temporal action detection on AVA and UCF101-24, and video scene graph classification on the recent Action Genome dataset -- and achieve state-of-the-art results on all three datasets.
Furthermore, we show quantitatively and qualitatively how our method is able to more effectively model relationships between relevant entities in the scene.

	\end{abstract}

\section{Introduction}

Deep learning has enabled rapid advances in many image understanding tasks such as image classification~\cite{he_cvpr_2016}, object detection~\cite{ren_neurips_2015} and semantic segmentation~\cite{chen_eccv_2018}.
However, progress on recent video understanding datasets such as AVA~\cite{gu_cvpr_2018} and Charades~\cite{sigurdsson_eccv_2016} has lagged behind in comparison.
Further progress in the video understanding tasks posed by these datasets would facilitate applications in autonomous vehicles, health monitoring and automated media analysis and production among others. %

\begin{figure*}[thb]
	\vspace{-0.5\baselineskip}
	\includegraphics[width=1\linewidth]{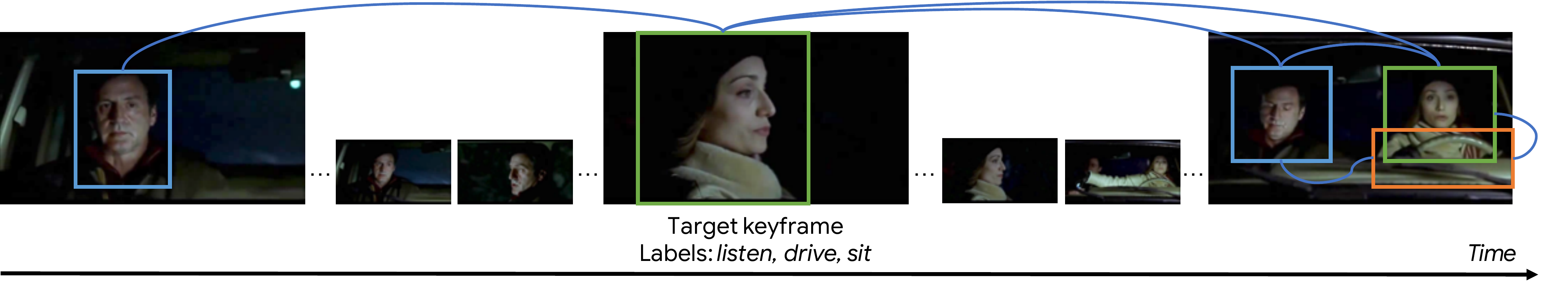}
	\caption{
	Understanding videos requires reasoning over long-term spatio-temporal interactions between actors, objects and the environment.
	The actions of the woman in the centre frame are ambiguous given the nearby frames which typical 3D CNN architectures consider.
	However, by considering her interactions with the man from nearby frames, we know she is ``listening'' to him.
	And the fact that she is later holding the steering wheel indicates she is ``driving'' in contrast to the man who is ``riding''.
	In this paper, we propose a spatio-temporal graph on which we perform message passing to explicitly model these spatio-temporal interactions.
	Example from the AVA dataset~\cite{gu_cvpr_2018}.
	}
	\label{fig:teaser}
	\vspace{-0.5\baselineskip}
\end{figure*}

A reason why video understanding is so challenging is because, as shown in Fig.~\ref{fig:teaser}, it requires understanding the interactions between actors, objects and other context in the scene.
Furthermore, these interactions are not always observable from a single frame, and thus require reasoning over long temporal intervals.
This is illustrated in Fig.~\ref{fig:teaser}, where understanding the actions of the person in the centre-frame is not possible from the target keyframe alone. %
In order to know that the woman is ``listening'', we need to consider the man who is speaking but no longer in the scene.
And to correctly infer that the woman is ``driving'' the car, rather than ``riding'' like the man, we must take into account that she is later holding the steering wheel. %

Video is a significantly higher-dimensional signal than single images, due to its additional temporal axis, and so we believe %
learning these unlabelled interactions directly from current datasets with large convolutional networks is not feasible. %
In this paper, we propose a structured graph neural network to explicitly model these spatio-temporal interactions.
We model actors and objects (explicitly with bounding boxes when we have supervision, and implicitly otherwise) as nodes in our spatio-temporal graph and perform message passing inference to directly model their relations.

Although a wide range of graph-structured models have been proposed for action recognition, we note that there has been no unifying formulation for these models.
As such, some works only model spatial relations between actors and objects~\cite{girdhar_cvpr_2019, sun_eccv_2018}, but not how these interactions evolve over time.
Other approaches model long-range temporal interactions~\cite{wu_cvpr_2019}, but do not capture spatial relations and are not trained end-to-end.
And whilst some methods do model spatio-temporal interactions of objects~\cite{baradel_eccv_2018,wang_eccv_2018}, their explicit representations of objects need additional supervision, and are not evaluated on spatio-temporal localisation tasks which requires detailed understanding and is necessary for analysing untrimmed videos.

Our graph network formulation based on the message-passing neural network~\cite{gilmer_icml_2017} abstraction, %
allows us to explicitly model interactions between actors, objects and the scene, and how these interactions evolve over time.
Our flexible model allows us to use explicit object representations from a pretrained Region Proposal Network (RPN)~\cite{ren_neurips_2015}, and/or implicitly from convolutional feature maps without additional supervision. %
Moreover, our general formulation allows us to interpret previous work~\cite{girdhar_cvpr_2019, sun_eccv_2018, wang_eccv_2018, wu_cvpr_2019, zhang_tokmakov_cvpr_2019} as special cases, and thus understand how different design choices in object representation, graph connectivity and message passing functions affect the model's performance.
We demonstrate our versatile model on two different tasks: spatio-temporal action detection on AVA~\cite{gu_cvpr_2018} and UCF101-24~\cite{soomro_arxiv_2012}, and video scene graph prediction on the recent Action Genome~\cite{ji_cvpr_2020} dataset.
Both of these tasks require modelling the spatio-temporal interactions between the actors and/or objects in the scene, and we show consistent improvements from using each component of our model and achieve state-of-the-art results on each dataset.
Furthermore, we observe that the largest improvements in AVA are indeed achieved on action classes involving human-to-human and human-to-object interactions, and visualisations of our network show that it is focusing on scene context that is intuitively relevant to its action classification.

\section{Related Work}

Modelling contextual relationships has a long history in scene understanding.
Relevant examples of early works in this area included modelling interactions between humans and objects~\cite{gupta_pami_2009, yao_cvpr_2010}, different objects~\cite{rabinovich_iccv_2007} and relationships between human actions and scene context~\cite{marszalek_cvpr_2009, gkioxari_iccv_2015}.
Furthermore, it has also been shown that human vision is reliant on context too~\cite{oliva_2007}.
In this paper, we consider video understanding tasks, specifically spatio-temporal action recognition and video scene graph parsing, which involve reasoning about interactions between actors, objects and their environment in both space and time.

Early work in action recognition used hand-crafted features to encode motion information~\cite{laptev_ijcv_2005, wang_ijcv_2013}.
Advances in deep learning first saw repurposing of 2D image convolutional neural networks (CNNs) for video as ``two stream'' networks~\cite{karpathy_cvpr_2014,simonyan_neurips_2014} followed by spatio-temporal 3D CNNs~\cite{carreira_cvpr_2017,feichtenhofer_neurips_2016,tran_cvpr_2018,zhou_eccv_2018}.
However, these architectures focus on extracting coarse, video-level features and are not suitable for learning the fine-grained relations depicted in Fig.~\ref{fig:teaser}.
Consequently, whilst initial approaches to spatio-temporal action detection involved extending 2D object detectors~\cite{liu_ssd_eccv_2016, ren_neurips_2015} temporally~\cite{kalogeiton_iccv_2017,singh_iccv_2017,arnab_eccv_2020, xu_star_arxiv_2020}, 
current leading methods~\cite{girdhar_cvpr_2019, wu_cvpr_2019, zhang_tokmakov_cvpr_2019} on the AVA dataset~\cite{gu_cvpr_2018} all explicitly model relationships, with approaches which we show can be interpreted as variants of graph neural networks.

Graph neural networks (GNNs) explicitly model relations between entities by modelling them as nodes in a directed or undirected graph~\cite{battaglia_arxiv_2018, scarselli_2008, kipf_iclr_2017}, which interact via a neighbourhood defined for each node.
The self-attention~\cite{vaswani_neurips_2017} and Non-local~\cite{wang_cvpr_2018} operators can also be thought of as GNNs~\cite{battaglia_arxiv_2018, gilmer_icml_2017} where each element in a feature map is a node, and all nodes are fully-connected to each other.
Such attention-based models have excelled at a number of natural language processing and computer vision tasks and have inspired many follow-up methods~\cite{hu_squeeze_excite_cvpr_2018, chen_a2_net_neurips_2018, chen_glore_cvpr_2019, wu_iclr_2019, zhang_dgmn_cvpr_2020}.

\begin{figure*}[thb]
	\vspace{-\baselineskip}
	\includegraphics[width=1\linewidth]{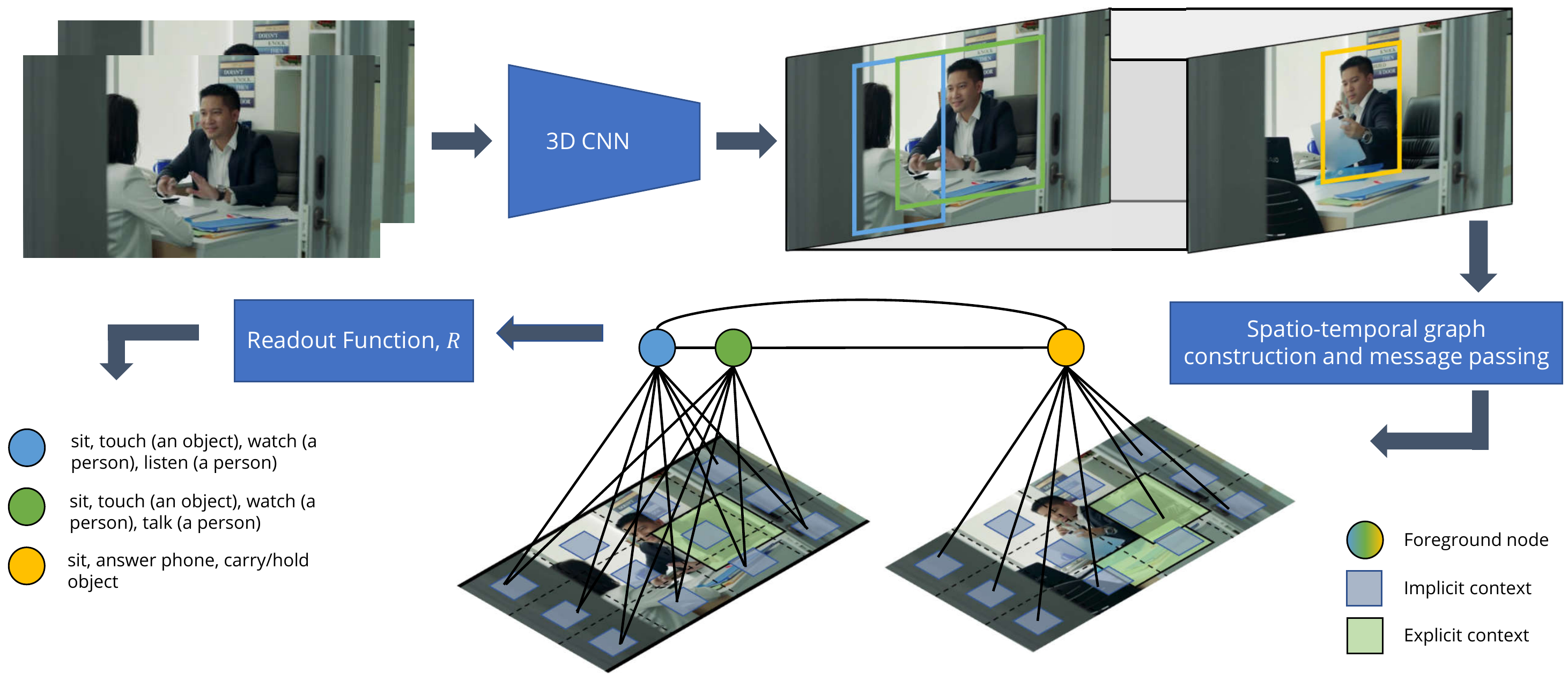}
	\caption{
	Overview of our method:
	We construct a spatio-temporal graph, and perform message-passing inference on it, to model interactions between actors, objects and their environment.
	Foreground nodes (circles) have readout functions associated with them for the task of interest (\ie for action recognition, the nodes represent person bounding boxes which are classified into actions).
	Context nodes (squares) model additional information, and can either be implicit, as cells of the original feature map, or explicit by ROI-pooling external region proposals (RPN~\cite{ren_neurips_2015} not shown for clarity).
	The initial state of each node is a spatio-temporal feature vector extracted from a 3D CNN.
	}
	\label{fig:method_diagram}
	\vspace{-0.5\baselineskip}
\end{figure*}

Many structured models have also recently been employed in video understanding.
However, there has previously been no coherent framework to unify these approaches.
Consequently, some works only model spatial relations between actors and objects, but not how these evolve over time~\cite{sun_eccv_2018, girdhar_cvpr_2019}.
And whilst LFB~\cite{wu_cvpr_2019} models long range temporal interactions, it does not capture spatial relationships within a keyframe.
Moreover, in order to model long-range interactions, \cite{wu_cvpr_2019} uses precomputed features and is thus not trained end-to-end.
Our proposed method, which is based on the message passing neural network (MPNN)~\cite{gilmer_icml_2017} framework, coherently models both spatial and temporal interactions. 
After describing our model in Sec.~\ref{sec:method}, we show how previous structured models for action detection~\cite{sun_eccv_2018, girdhar_cvpr_2019,wang_eccv_2018, wu_cvpr_2019, zhang_tokmakov_cvpr_2019} can be thought of as special cases of our model.
Furthermore, our flexible model can reason about objects both when we have explicit supervision for them, and when we do not.
Prior work in video understanding, which have proposed less generic models, have either assumed the case of explicit object supervision~\cite{baradel_eccv_2018, wang_eccv_2018} or not~\cite{sun_eccv_2018, girdhar_cvpr_2019, zhang_tokmakov_cvpr_2019}, but have not considered the scenario when both options are available.
Furthermore, our general formulation allows us to ablate graph modelling design choices, such as the object representation, message passing functions and temporal context, in a manner not possible with these previous, more specific approaches.
Moreover, we note that some approaches which have employed spatio-temporal graphs in video have only considered frame-to-frame interactions~\cite{jain_cvpr_2016, wang_eccv_2018, baradel_eccv_2018, ma_cvpr_2018, materzynska_cvpr_2020, qi_eccv_2018}, rather than long-range relations, and have not evaluated on spatio-temporal localisation which requires more detailed understanding and is essential for analysing untrimmed videos.

We also note that scene graph parsing~\cite{johnson_cvpr_2015, krishna_ijcv_2017} is another task that evaluates a model's ability to reason about the interactions between different objects by representing objects as nodes and relationships as edges in a graph.
Although the task was originally posed for single images~\cite{krishna_ijcv_2017}, the recent Action Genome~\cite{ji_cvpr_2020} dataset extends this task to video by adding annotations to Charades~\cite{sigurdsson_eccv_2016}.
While GNN-based approaches have also been used in scene graph parsing~\cite{dai_drn_cvpr_2017,li_vipcnn_cvpr_2017,xu_cvpr_2017,yang_graphrcnn_eccv_2018} for single images, to our knowledge, they have not been applied to model temporal relations in video.
Moreover, the same model has not also been demonstrated on spatio-temporal action recognition like our method.

\section{Proposed Approach} %
\label{sec:method}

Our model aims to build a structured representation of a video by representing it as a graph of actors, objects and contextual elements in the scene, as shown in Fig.~\ref{fig:method_diagram}.
This structured representation is then used to perform tasks which require understanding the interactions between the elements in the graph, such as action recognition and scene graph prediction. 
Note that we do not assume we have annotations for relevant scene context.

Our approach is based on Message Passing Neural Networks (MPNN)~\cite{gilmer_icml_2017}, as it is a flexible framework that generalises many previous graph neural network algorithms~\cite{battaglia_neurips_2016,battaglia_arxiv_2018,kipf_iclr_2017,velivckovic_iclr_2018} %
We review this approach, and describe how we adapt it for video understanding in Sec.~\ref{sec:method_mpnn}, before detailing our model in Sec.~\ref{sec:method_spatial_model} through~\ref{sec:method_readout_function}.
Finally, we discuss how previous structured models for video understanding~\cite{girdhar_cvpr_2019, sun_eccv_2018, wang_eccv_2018,wu_cvpr_2019, zhang_tokmakov_cvpr_2019} can be regarded as specific instantiations of our model in Sec.~\ref{sec:method_discussion}.

\subsection{Message Passing Neural Networks (MPNNs)}
\label{sec:method_mpnn}

MPNNs operate on a directed %
or undirected graph, $\mathcal{G}$ consisting of nodes, $v \in \mathcal{V}$, and a neighbourhood for each node, $\mathcal{N}_v$, that defines the graph's connectivity.
For video models we distinguish between the spatial, $\mathcal{S}_v$, and temporal, $\mathcal{T}_v$, neighbourhoods for node $v$ ($\mathcal{N}_v$ = $\mathcal{S}_v \cup \mathcal{T}_v$).
Each node, $v$, is associated with a latent state, $h_v$.
Inference in this model consists of a message passing phase, and a final readout phase.
In the message passing phase, messages for each node, $m_v$, are first computed by applying spatial and temporal message passing functions, $M_s$ and $M_t$ respectively, 
to all nodes in its neighbourhood as described in Eq.~\eqref{eq:mpnn_message_passing}.
An update function, $U$, then aggregates the received messages to update the latent state, $h_v$,
\begin{align}
m_v^{i + 1} &= \sum_{w \in \mathcal{S}_v}{M_s(h_v^i, h_w^i ; \theta_s^i)} + \sum_{u \in \mathcal{T}_v}{M_t(h_v^i, h_u^i ; \theta_t^i)} \label{eq:mpnn_message_passing}\\
h_v^{i + 1} &= U(m_v^{i + 1}, h_v^i), \label{eq:mpnn_update_function}
\end{align}
where $\theta$ denotes learnable function parameters.
Intuitively, the state of a node, $h_v$, is updated by aggregating the messages passed from its neighbours.
Finally, after $I \geq 1$ iterations of message passing, a readout function, $R$, uses the updated node features for the classification task of interest
\begin{equation}
y = R(\{h_v^i\} | v \in \mathcal{G}). \label{eq:mpnn_readout_function}
\end{equation}

As illustrated in Fig.~\ref{fig:method_diagram}, our graph consists of a set of ``Foreground'' nodes, $\mathcal{F} = \{f_1, f_2, \ldots, f_n\}$ and ``Context'' nodes, $\mathcal{C} = \{c_1, c_2, \ldots, c_m\}$ where $n$ and $m$ vary for each video.
The ``Foreground'' nodes, $\mathcal{F}$, have readout and loss functions associated with them, and
correspond to the objects that will subsequently be classified.
For the task of action recognition, $\mathcal{F}$ corresponds to bounding boxes of each actor in the keyframe.
Whilst for scene graph prediction, $\mathcal{F}$ refers to bounding boxes for all potential objects of interest.
The ``Context'' nodes, $\mathcal{C}$, capture additional information extracted from the scene for relational reasoning.
The representations of these nodes, and their spatial connectivity, are detailed next.

\subsection{Spatial Model}
\label{sec:method_spatial_model}

The spatial connections in our graph model relationships between actors, objects and scene context in the same frame.
For example, recognising an action such as ``give object to a person'' in the AVA dataset~\cite{gu_cvpr_2018} requires understanding both people involved in the action and also the object being transferred.

To model such interactions, we first represent the foreground nodes of our model, $\mathcal{F}$, by extracting convolutional features from the last layer of the network, $\mathbf{X} \in \mathbb{R}^{t \times h \times w \times c}$, and using ROI-Align~\cite{he_iccv_2017} followed by spatio-temporal pooling and a linear projection to obtain $f_i \in \mathbb{R}^{d}$.

We model scene context by considering the features from each spatial position in the feature map, $\mathbf{X}$, as a contextual node, $c_i$, in our graph.
Note that these features are projected to $\mathbb{R}^{d}$.
A similar representation was used by \cite{sun_eccv_2018}, which we refer to as an \emph{implicit} object model as it enables the network to encode information about the scene and relevant objects without any extra supervision.
This approach is also known as ``grid features'' in visual question answering~\cite{battaglia_neurips_2016,jiang_cvpr_2020}.

It is also possible to augment our set of contextual nodes with an \emph{explicit} object representation by computing class-agnostic object proposals with a Region Proposal Network (RPN)~\cite{ren_neurips_2015}.
We use an RPN pretrained on the OpenImages dataset~\cite{kuznetsova_ijcv_2020} and obtain a $d$-dimensional feature from each proposal by using ROI-Align and a linear projection as for the foreground nodes.
A similar idea of using explicit object representations in video has also been employed by \cite{baradel_eccv_2018} and \cite{wang_eccv_2018}, though not for spatio-temporal action recognition.

Messages are then computed and passed to foreground nodes from both foreground and contextual nodes in the graph.
We only update the foreground nodes during message passing, as these are the nodes which are subsequently classified in the final Readout phase. 
Concretely, the spatial neighbourhood for each foreground node in our graph is
\begin{equation} 
\mathcal{S}_v =\mathcal{F} \cup \mathcal{C} \; \forall v \in \mathcal{F}.
\label{eq:spatial_neighbourhood}
\end{equation}

\subsection{Temporal Model}
\label{sec:method_temporal_messages}
\begin{figure}
	\includegraphics[width=0.99\linewidth]{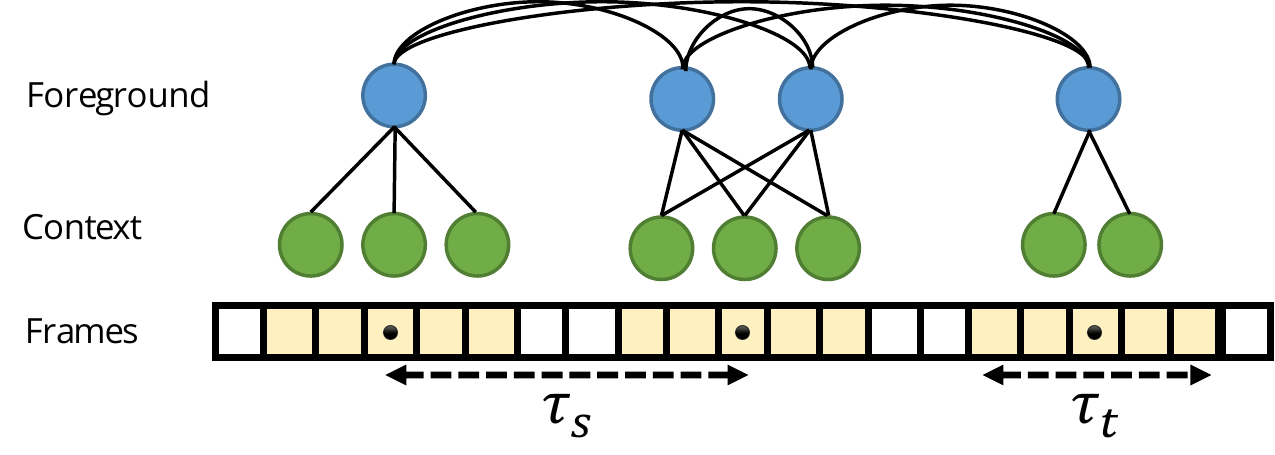}
	\caption{
	Illustration of temporal connectivity.
	Foreground (blue) and context (green) nodes have initial states computed from spatio-temporal features over $\tau_t$ frames (shaded in yellow), centred on a keyframe (denoted by a black circle).
	$\tau_s$ controls the distance between selected keyframes; $\tau_c$ is the total number of keyframes considered.
	Here, $\tau_c = 3$, $\tau_s =7$ frames and $\tau_t$ = 5 frames. 
	}
	\vspace{-\baselineskip}
	\label{fig:temporal_connections}
\end{figure}

We also include temporal connections in our graph to model long-range interactions between actors and objects.
As shown in Fig.~\ref{fig:teaser}, understanding actions often requires reasoning about actors who are no longer visible in the current frame, thus requiring large temporal contexts.

We model these temporal interactions by connecting foreground nodes in keyframe $t$ with all other foreground nodes in neighbouring keyframes $t' \in \mathcal{T}$.
Concretely, we define the temporal context, $\tau_c$, and temporal stride, $\tau_s$, hyperparameters.
As shown in Fig.~\ref{fig:temporal_connections}, $\tau_c$ is the total number of number of keyframes in the video which we consider in our temporal graph.
$\tau_s$ is the sampling rate at which we select keyframes, as $\tau_s \ge 1$ allows us to consider a wider temporal interval in a more computationally efficient manner.
This is necessary to train the entire model end-to-end.
Moreover, as each foreground feature node in the graph, $f_i$, is itself a spatio-temporal feature computed over a period of $\tau_t$ by a 3D CNN, selecting adjacent keyframes (and effectively setting $\tau_s = 1$ keyframe) could result in redundant information being captured by the temporal connections in the graph. %
Note that our definition of keyframe follows common datasets~\cite{gu_cvpr_2018, ji_cvpr_2020} as a frame we aim to classify given surrounding temporal context.

More formally, we can describe the neighbourhood of each foreground node, $v$, for temporal message passing as
\begin{equation}
\mathcal{T}_v = \bigcup_{\ceil{t=-\tau_c / 2}}^{\floor{\tau_c / 2}}{\mathcal{F}^{t \cdot \tau_s}}.
\end{equation}
Here, we use the superscript to denote the temporal index, and without loss of generality consider $t = 0$ to be the centre keyframe, meaning that negative time indices correspond to frames in the past.
We set $\tau_c$ to be an odd, positive integer to employ an equal-sized temporal window on either side of the centre keyframe.

We first perform spatial message passing, before passing messages temporally.
This allows information from the context nodes in frame $t'$, $c_j^{t'}$, to efficiently propagate to a foreground node $f_i^t$ (where $t \neq t'$) via $f_j^{t'}$, as the foreground nodes are fully-connected temporally.

The following section now describes the messages that are passed along the graph described above.

\subsection{Message Passing Functions}
\label{sec:method_message_passing_function}

We first observe that the Non-local operator or self-attention~\cite{vaswani_neurips_2017,wang_cvpr_2018} can be considered as a message passing function in a fully-connected graph, where each node, $h_v$, is an element in the input feature map, $\mathbf{H}$, and the neighbourhood comprises of all other feature elements.
And since self-attention is employed with a residual connection~\cite{wang_cvpr_2018,vaswani_neurips_2017} and layer normalisation~\cite{ba_arxiv_2016,vaswani_neurips_2017}, the update function of Non-local when viewed as an MPNN is
\begin{equation}
U = \text{LN}\left(h_v + \text{Self-Attention}\left(\mathbf{H}\right)\right).
\label{eq:update_func}
\end{equation}
Similar analysis~\cite{battaglia_arxiv_2018, gilmer_icml_2017} has shown that Graph Attention Networks (GAT)~\cite{velivckovic_iclr_2018}, Relational Networks~\cite{santoro_neurips_2017} and many other graph neural networks~\cite{scarselli_2008, kipf_iclr_2017,li_ggsnn_iclr_2016,battaglia_neurips_2016} can also be interpreted as MPNNs.
In this paper, we consider Non-local~\cite{wang_cvpr_2018} and Graph Attention (GAT)~\cite{velivckovic_iclr_2018} as message passing functions within the update function, Eq.~\eqref{eq:update_func}.

\vspace{\paravspace}
\paragraph{Non-local} 
We modify Non-local~\cite{wang_cvpr_2018} to pass messages from all Foreground and Context nodes to only Foreground nodes (Eq.~\ref{eq:spatial_neighbourhood}), as these are the nodes which are used for the final classification in the Readout phase,
\begin{gather}
\mathbf{M} = \text{Softmax}\left(\frac{\mathbf{Q}\mathbf{K}^\top}{\sqrt{d}}\right)\mathbf{V} \\
\mathbf{Q} = \mathbf{A}\mathbf{W}_q \qquad \mathbf{K} = \mathbf{[A || C]}\mathbf{W}_k \qquad \mathbf{V} = \mathbf{[A || C]}\mathbf{W}_v. \nonumber
\end{gather}
Here, $\mathbf{A} \in \mathbb{R}^{n \times d}$ and $\mathbf{C} \in \mathbb{R}^{m \times d}$ are matrices where each row is a foreground %
and context feature node respectively, $\mathbf{M}$ is the $\mathbb{R}^{n \times d}$ matrix of messages received by each node in $\mathbf{A}$, $\mathbf{[A || C]}$ denotes the concatenation of these matrices and $\mathbf{W}_q$, $\mathbf{W}_k$ and $\mathbf{W}_v$ are learnable $d \times d$ projection matrices.

\vspace{\paravspace}
\paragraph{Graph attention (GAT)}
The graph attention ~\cite{velivckovic_iclr_2018} message for a node $v$ is computed as
\begin{align}
m_v &= \sigma\left(\sum_{j \in \mathcal{N}_v}\alpha_{ij}\mathbf{W}_ah_j\right) \\
\alpha_{ij} &= \text{Softmax}\left(\sigma\left(w_b^{\top}[h_i || h_j]\right)\right) \label{eq:att_weights}
\end{align}
where $\sigma$ is a ReLU non-linearity and $\mathbf{W}_a$ and $w_b$ are a learnable matrix and vector respectively.

\vspace{\paravspace}
\paragraph{Parallel messages}

It is also possible to compute multiple incoming messages in parallel for a node, $h_v$.
When using multiple Non-local or Graph Attention functions, this corresponds to multi-headed attention~\cite{vaswani_neurips_2017, velivckovic_iclr_2018}.
By viewing Non-local and GAT as message passing functions, we can also aggregate messages from a combination of these two.
In these cases, we aggregate the messages using an attention-weighted convex combination, as performed in Eq.~\eqref{eq:att_weights}.

\subsection{Readout Function}
\label{sec:method_readout_function}

After $I \geq 1$ iterations of message passing, a readout function is applied on the Foreground nodes to obtain the final predictions.
For action detection, the readout function is a linear classifier operating on each element of $\mathcal{F}$, where each foreground node, $f_i$, corresponds to the features of an actor in the keyframe.
For scene graph prediction, the readout function consists of two classifiers: %
The first linear classifier predicts the object-class label of each foreground node.
The second is a function of each pair of foreground nodes, and predicts the relationship label between them.

\subsection{Discussion}
\label{sec:method_discussion}
We note that many previous structured models for video understanding can be considered as special cases of our proposed MPNN framework:

Girdhar~\etal~\cite{girdhar_cvpr_2019} only consider a spatial model, \ie $\mathcal{N} = \mathcal{S}$, using implicit objects and Non-local~\cite{wang_eccv_2018} for message passing.
ACRN~\cite{sun_eccv_2018} has the same graph structure, but uses Relational Networks~\cite{santoro_neurips_2017} for message passing instead.
LFB~\cite{wu_cvpr_2019}, in contrast, considers only a temporal model, \ie $\mathcal{N} = \mathcal{T}$, using Non-local as the message passing function in a graph that is fully-connected in time.
However,~\cite{wu_cvpr_2019} do not consider a spatial model to capture interactions between actors in the keyframe.
Zhang~\etal~\cite{zhang_tokmakov_cvpr_2019}, on the other hand, model both temporal and spatial connections.
However, they effectively model three separate graphs: the first models actors in short, 3-second tubelets using GCN~\cite{kipf_iclr_2017} for message passing.
The other two graphs model actor-actor and actor-object relations using a message passing method similar to GAT~\cite{velivckovic_iclr_2018}, but using a hand-defined weighting function rather than a learned one as in GAT (Eq.~\ref{eq:att_weights}).
Wang~\etal~\cite{wang_eccv_2018} also model a spatio-temporal graph, using GCN for message passing, and an explicit object representation in the spatial model.
However, their temporal connections are only among adjacent frames, which does not allow information propagation between all frames in long sequences, in contrast to our model which is fully-connected temporally.
With our unified framework, we study the effect of graph modelling design choices and show how we outperform previous work next in Sec.~\ref{sec:experiments}.

\section{Experiments}
\label{sec:experiments}

\begin{table*}[tb]
	\captionsetup{skip=3pt}
	\caption{Ablation study on Action Genome using a 3D ResNet 50 backbone. We report the effect of (a) different message passing functions, (b) temporal connections in the graph ($\tau_c = 1, \tau_s = 1$ corresponds to only spatial message passing) and (c) iterations of message passing.}
	\begin{subtable}[t]{.41\linewidth}
		\centering
		\captionsetup{font=footnotesize, skip=1pt}
		\caption{Spatial message passing functions}
		\renewcommand*{\arraystretch}{1.19}
		\scriptsize{
			\begin{tabular}{lcccc}
				\toprule
				& \multicolumn{2}{c}{SGCls} & \multicolumn{2}{c}{PredCls} \\ \cmidrule(lr){2-3} \cmidrule(lr){4-5}
				& R@20        & R@50        & R@10         	& R@20         \\ \midrule
				Baseline 			&   48.9          &  51.3           &    78.7          	&       93.8       \\ %
				Non-local in backbone	&   49.1      &  51.4           &  78.8            	&   93.9           \\  %
				\midrule
				Non-local 			&	50.4		  & 	52.6		&	79.3		   	& 	94.2		   \\ %
				GAT				  	&	51.1		  & 	53.2		&	\textbf{79.7}		   	&	\textbf{94.4}		   \\ %
				GAT + Non-local 	&	\textbf{51.3}		  &	\textbf{53.4}			&	79.4		   	& 94.3			   \\  %
				\bottomrule
			\end{tabular}
		}
		\label{tab:ablation_spatial_message_passing}
	\end{subtable}
	\hfill
	\begin{subtable}[t]{.38\linewidth}
		\centering
		\captionsetup{font=footnotesize, skip=1pt}
		\caption{Temporal message passing}
		\scriptsize{
			\begin{tabularx}{1\linewidth}{YYYYYY}
				\toprule
				
				\multicolumn{2}{c}{Temporal parameters} & \multicolumn{2}{c}{SGCls} & \multicolumn{2}{c}{PredCls} \\ \cmidrule(lr){1-2} \cmidrule(lr){3-4} \cmidrule(lr){5-6}
				$\tau_c$ & $\tau_s$					& R@20        & R@50        & R@10         	& R@20         \\ \midrule
				1 & 1 			&   51.1		  & 	53.2		&	\textbf{79.7}		   	&	\textbf{94.4}		   \\ 
				3 & 2	&   52.9      &  55.0           &  79.4            	&   94.2           \\  
				3 & 5				  	&	53.3		  & 	55.5		&	79.4		   	&	94.2		   \\ 
				3 & 7 	& 	53.5	& 55.7	& 79.3 	& 94.2			\\  %
				5 & 2 	&	53.4		  &	55.5			&	79.4		   	& 94.2			   \\ 
				5 & 5 	&	\textbf{53.8}		  &	\textbf{56.0}			&	79.3		   	& 94.2			   \\  
				5 & 7 	&	53.6		  &	55.8			&	79.3		   	& 94.2			   \\  
				\bottomrule
			\end{tabularx}
		}
		\label{tab:ablation_temporal_model}
	\end{subtable}
	\hfill
	\begin{subtable}[t]{.20\linewidth}
		\centering
		\captionsetup{font=footnotesize, skip=1pt}
		\caption{Message passing iterations}
		\renewcommand*{\arraystretch}{1.2}
		\scriptsize{
			\begin{tabular}{cc}
				\toprule
				Iterations & R@20 \\
				\midrule
				1 & 51.1 \\
				2 & 51.6 \\
				3 & 51.6 \\
				5 & 51.8 \\
				\bottomrule
			\end{tabular}
		}
		\label{tab:ablation_iterations}
	\end{subtable}
	\vspace{-1\baselineskip}
\end{table*}

We evaluate our method's ability to model spatio-temporal interactions with experiments on spatio-temporal action detection and video scene graph classification.

\vspace{\paravspace}
\paragraph{Spatio-temporal action recognition}
We evaluate on AVA~\cite{gu_cvpr_2018}, the largest dataset for this task consisting of 15-minute video clips obtained from movies, and UCF101-24~\cite{soomro_arxiv_2012}, the previous standard benchmark for this task.
AVA is labelled with atomic actions, where one person is typically performing multiple actions simultaneously, whilst actors perform only a single high-level action in UCF101-24.
We follow standard protocol and evaluate using the Frame AP at an IoU threshold of 0.5 on both datasets.
For AVA, we use v2.2 annotations for ablations, and either v2.1 or v2.2 annotations for fair comparisons with prior work.

\vspace{\paravspace}
\paragraph{Video scene graph prediction}
We evaluate on the recent Action Genome dataset~\cite{ji_cvpr_2020}, which adds scene graph annotations to  Charades~\cite{sigurdsson_eccv_2016}, on two scene graph tasks: scene graph classification (SGCls) and predicate classification (PredCls).
Both tasks are evaluated using the standard Recall@K metric~\cite{lu_vrd_eccv_2016,ji_cvpr_2020} (R@K) which measures the fraction of ground truth relationship triplets (subject-predicate-object) that appear in the top K scoring predicted triplets.
In SGCls, ground truth bounding box co-ordinates are given, and the aim is to predict their object classes, as well as relationship labels between pairs of objects.
PredCls is simpler, as both bounding box co-ordinates and object classes are given, and only the relationship label must be predicted.

\vspace{\paravspace}
\paragraph{Implementation details}

We use the public implementation of SlowFast~\cite{feichtenhofer_iccv_2019} as our baseline with the 3D ResNet 50 or ResNet 101 backbones~\cite{he_cvpr_2016}, as it is the current state-of-the-art.
The network is similar to Fast-RCNN~\cite{girshick_iccv_2015} as it uses external region proposals to extract features from the last feature map of $\text{res}_5$ using ROI-Align~\cite{he_iccv_2017}.
These features are then spatio-temporally pooled and classified.

In our graph model, we initialise the internal states, $h_i$, of our Foreground nodes using these ROI-Aligned  $\text{res}_5$ features.
For our action detection experiments, the Foreground nodes in our graph correspond to bounding boxes of actors, and we use the same person detector as \cite{feichtenhofer_iccv_2019, wu_cvpr_2019} for our actor region proposals.
For scene graph experiments, we use ground truth boxes of people and objects as our Foreground nodes as we evaluate scene graph classification (SGCls) and predicate classification (PredCls).%

Unless otherwise specified, we use SlowFast $8 \times 8$ which corresponds to 32 input frames where the video is subsampled by a factor of 2.
This means that each feature node, $h_i$, in the graph aggregates $\tau_t = 2.1$ seconds of temporal information on AVA as the videos are sampled at 30 frames per second (fps).
As keyframes in AVA are defined at 1 fps, it means that we need to set $\tau_s \geq 2$ keyframes for the temporal information captured by temporally-adjacent nodes in the graph to not overlap.
For Action Genome, the Charades videos are sampled at 24 fps and the keyframes are on average 0.85 seconds, or 20.5 frames apart.

We train our network for 20 epochs using synchronous SGD on 8 GPUs and a total batch size of 64, 
initialising from a Kinetics-400~\cite{kay_arxiv_2017} pretrained model.
The baseline model in all our experiments is SlowFast without any graph module trained in an identical manner.
When training spatio-temporal graph models, we finetune for 10 epochs from a model trained with only a spatial graph and reduce the batch size by a factor of $\tau_c$.
We also ``freeze'' batch normalisation statistics as they have high variance for small batches~\cite{he_iccv_2017}. %
Full training details are in the appendix.

\subsection{Experiments on video scene graph prediction}

\paragraph{Spatial message passing}

We first ablate only the spatial component (Sec.~\ref{sec:method_spatial_model}) of our model in Tab.~\ref{tab:ablation_spatial_message_passing}.
Graph Attention (GAT) ~\cite{velivckovic_iclr_2018} performs slightly better than Non-local~\cite{vaswani_neurips_2017, wang_cvpr_2018} as a message passing function, and we obtain a further small improvement by combining the two methods in parallel (Sec.~\ref{sec:method_message_passing_function}).
Overall, we improve upon our baseline model, which is SlowFast based on ResNet 50 without any graph modelling, by 2.4 points for R@20 on SGCls.

Another baseline is to insert a Non-Local layer~\cite{wang_cvpr_2018} after the final $\text{res}_5$ layer and before the ROI-Align layer, since Non-local can also be viewed as a graph network (Sec.~\ref{sec:method_message_passing_function}).
In this case, the Foreground and Context nodes are not explicitly modelled as in our method and the overall performance is 2.2 points less than our method for the R@20 of SGCls.
This suggests that explicit modelling of the Foreground nodes which are subsequently classified is important to performance.
We note, however, that Non-Local~\cite{wang_cvpr_2018} is typically employed earlier in a network (\ie in $\text{res}_3$ or $\text{res}_4$~\cite{wang_cvpr_2018, wu_cvpr_2019,feichtenhofer_iccv_2019}) and can thus be seen as a complementary method to improve features learned by the network.

\vspace{\paravspace}
\paragraph{Temporal message passing}

Table~\ref{tab:ablation_temporal_model} adds temporal connections (Sec.~\ref{sec:method_temporal_messages}) to our model, using GAT for message passing, as it outperformed Non-local in Tab.~\ref{tab:ablation_spatial_message_passing}. %
We observe consistent improvements for a wide range of temporal contexts, $\tau_c$, and strides, $\tau_s$, showing the utility of modelling temporal dynamics.
We improve over a graph with only spatial connections by as much as 2.7 points for R@20.

For a temporal model, another baseline to consider is to simply increase the temporal information available to the spatial-only graph model by increasing the number of frames at the input.
When tripling the number of input frames, the R@20 is 52.4, less than all variants of our temporal model with $\tau_c = 3$.
Similarly, increasing input frames by a factor of 5, the R@20 is 52.8, less than all our temporal models with $\tau_c = 5$.
This improvement demonstrates the benefits of message passing on an explicit temporal graph.

\vspace{\paravspace}
\paragraph{Iterations of message passing}
Table~\ref{tab:ablation_iterations} shows that the model's performance plateaus after 2 iterations of message passing, where parameters are not tied across iterations.

\begin{table}[tb]
	\centering
	\caption{Comparison to existing methods on the Action Genome dataset~\cite{ji_cvpr_2020}. Previous methods reported by~\cite{ji_cvpr_2020}.
	}
	\scalebox{0.9}{
		\begin{tabular}{lcccc}
			\toprule
			 & \multicolumn{2}{c}{SGCls} & \multicolumn{2}{c}{PredCls} \\ \cmidrule(lr){2-3} \cmidrule(lr){4-5}
								& R@20        & R@50        & R@10         	& R@20         \\ \midrule
			MSDN~\cite{li_iccv_2017}			&   44.0          &  47.2           &    --          	&       --       \\ 
			IMP~\cite{xu_cvpr_2017}	&   44.1      &  47.4           &  --            	&   --           \\  
			RelDN~\cite{zhang_gcl_cvpr_2019} 			&	46.7		  & 	49.4		&	--		   	& 	--		   \\ 
			\midrule
			SlowFast (ResNet 50)				  	&	48.9		  & 	51.3		&	78.7		   	&	93.8		   \\ %
			Ours (ResNet 50) 	&	\textbf{53.8}		  &	\textbf{56.0}			&	\textbf{79.3}		   	& \textbf{94.2}			   \\  %
			\bottomrule
		\end{tabular}
	}
	\label{tab:action_genome_sota_comparison}
\end{table}

\begin{figure}[tb]
	\vspace{-0\baselineskip}
	\centering
	\renewcommand{\arraystretch}{0.2}  %
	\setlength{\tabcolsep}{6pt} %
	
	\def \imageheight {1.8cm}
	
	\begin{tabular}{cc}
		\includegraphics[width=0.45\linewidth]{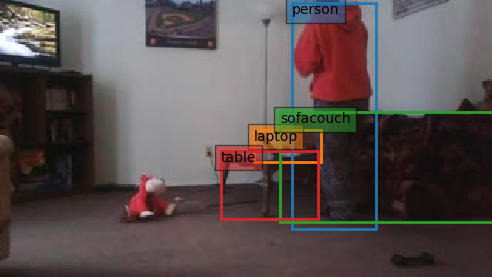}
		&
		\includegraphics[width=0.45\linewidth]{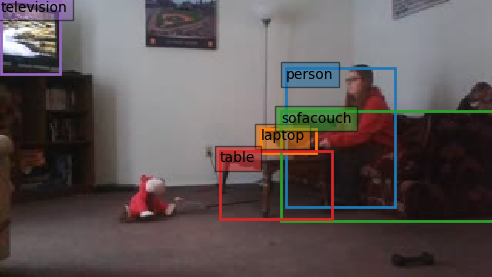}
		\\
		\includegraphics[height=3.3cm]{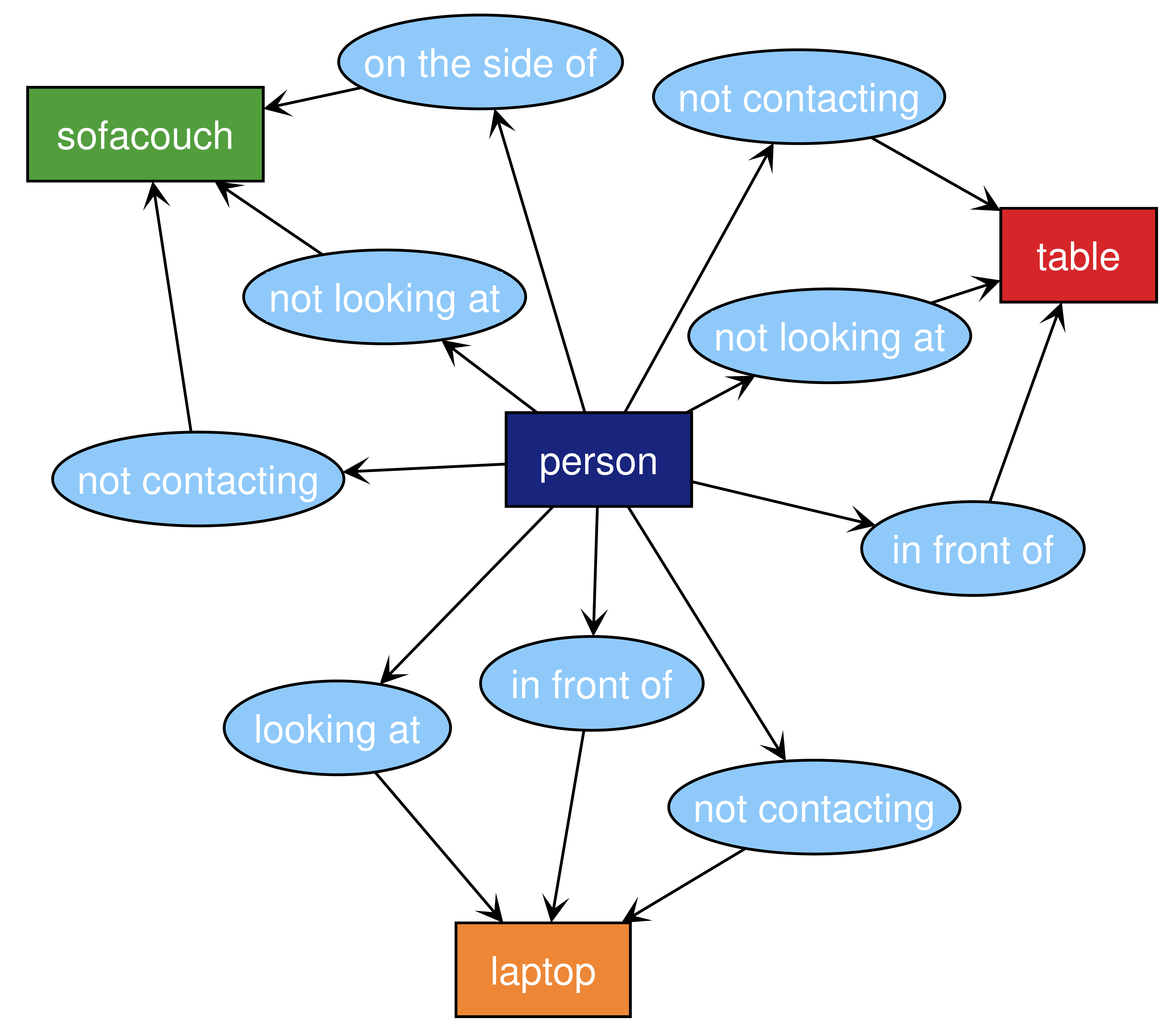}
		&
		\includegraphics[height=3.3cm]{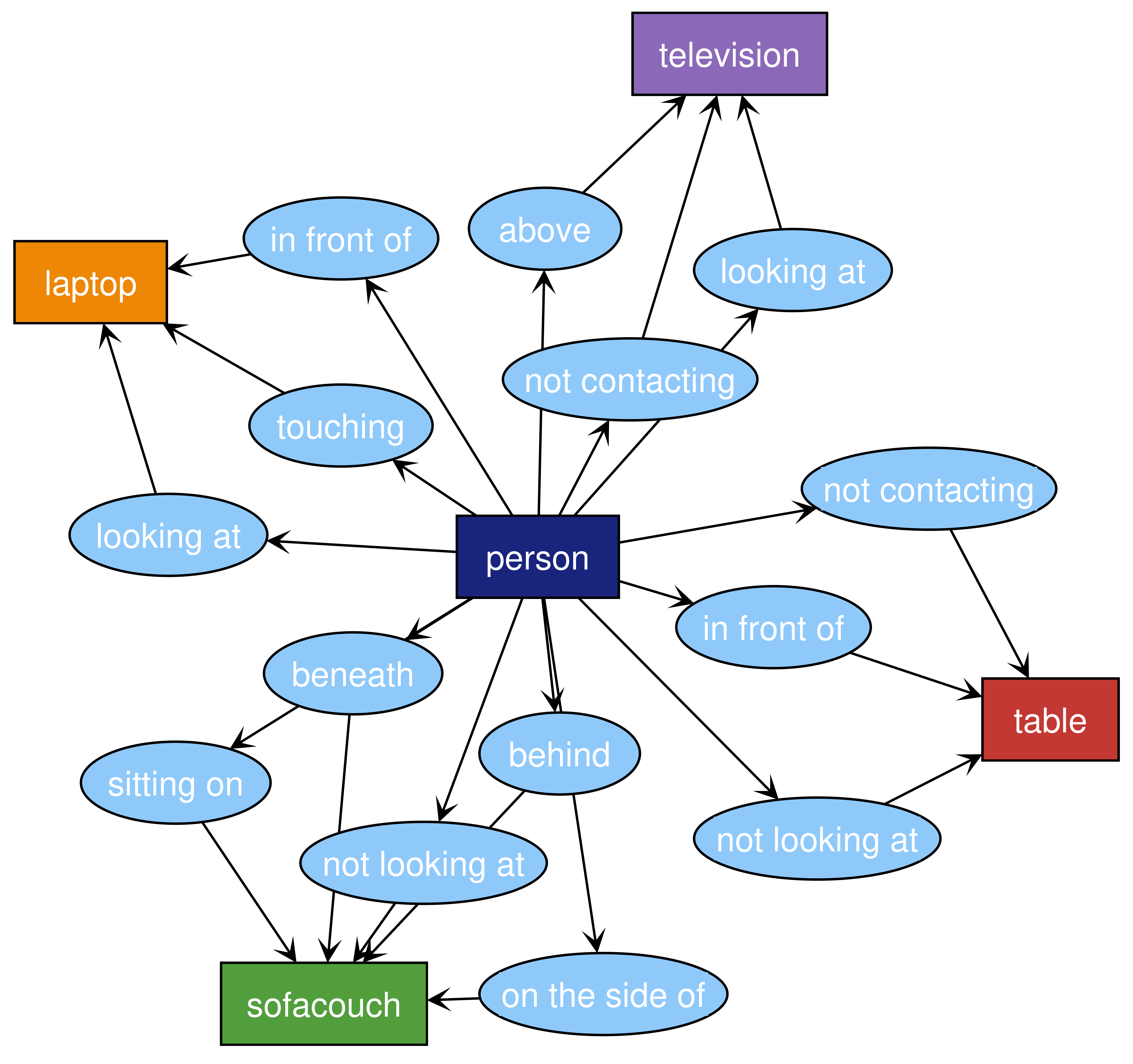}
	\end{tabular}
	\caption{
		Example scene graph predictions for consecutive keyframes of Action Genome~\cite{ji_cvpr_2020}. Classified bounding boxes are on top, and the corresponding scene graph (objects are shown as coloured rectangles, and relations with light-blue ovals) below.
	}
	\label{fig:action_genome_example}
	\renewcommand{\arraystretch}{1}  %
	\setlength{\tabcolsep}{6pt}  %
	\vspace{-1\baselineskip}
\end{figure}

\begin{table}[tb]
	\centering
	\caption{
		Analysis of message passing neighbourhood on AVA using SlowFast with a ResNet-50 backbone as the baseline. 
		By varying the neighbourhood, we study the effect of object representations (implicit, explicit or none) and temporal connections.
		We report the Frame mAP on the three types of action classes in AVA.
	}
	\scalebox{0.83}{
		\begin{tabularx}{1.2\linewidth}{l YYYY}
			\toprule
			Message passing neighbourhood & Pose & Human-Human & Human-Object & All \\ \midrule
			SlowFast baseline (none) &  43.1    &   25.2          &   17.4           &    24.8     \\
			\midrule 
			Actors only	    		&   43.2   &  27.0           &    17.8          &   25.6      \\
			Implicit objects only	&  43.4    &   26.7          &     18.0         &   25.7      \\
			Explicit objects only 	   &  43.0    &  26.7           &   17.8           &   25.5      \\
			Actor + Implicit  &  43.4    &    26.8         &     18.3         &   25.9      \\
			Actor + Implicit + Explicit  &  43.7    &   27.0          &   18.4           &   26.1      \\
			\midrule
			Spatio-temporal & 43.8 & 27.5 & 19.9 & 27.0 \\
			\bottomrule
		\end{tabularx}
	}
	\vspace{-\baselineskip}
	\label{tab:ablation_ava_message_passing}
\end{table}

\vspace{\paravspace}
\paragraph{Comparison to state-of-the-art}
Finally, we compare to prior works in Tab.~\ref{tab:action_genome_sota_comparison} which are single-image models evaluated by~\cite{ji_cvpr_2020}.
Our SlowFast, ResNet 50 3D baseline outperforms these, showing the importance of using spatio-temporal features for this task.
As Action Genome contains videos of humans acting to scripts, there is temporal structure in the interactions between actors and objects in the scene (Fig.~\ref{fig:action_genome_example}).
Our final spatio-temporal graph structured model improves substantially upon this baseline by 4.9 and 4.7 points for the R@20 and R@50 for SGCls respectively.
Our improvements over the baseline for PredCls are less, as this task is easier and the performance is saturated.

\subsection{Experiments on action recognition}

\paragraph{Graph structure and object representation}
Table~\ref{tab:ablation_ava_message_passing} compares the effect of changing the neighbourhood that passes messages to each foreground node that is subsequently classified.
We report the performance across the three types of action categories in AVA~\cite{gu_cvpr_2018} -- Pose, Human-Human and Human-Object -- to show the effect that different object representations (implicit, explicit or neither as described in Sec.~\ref{sec:method_spatial_model}) and also temporal connections have.
Our SlowFast baseline does not model any explicit graph and hence effectively performs no message passing.
For our AVA experiments, Foreground nodes in the graph correspond to bounding boxes of the actors in the scene, using the same person detections as \cite{wu_cvpr_2019,feichtenhofer_iccv_2019}.
Thus, in this section, ``Foreground'' and ``actor'' nodes are used interchangeably.

\begin{figure*}[tb]
	\vspace{-0.5\baselineskip}
	\centering
	\renewcommand{\arraystretch}{0.8}  %
	\setlength{\tabcolsep}{1pt} %
	
	\def \imageheight {1.86cm}
	\def \trimleft {0.2cm}

	\begin{tabular}{ccccccl}
		\includegraphics[height=\imageheight, trim={2cm 0 0 0}, clip]{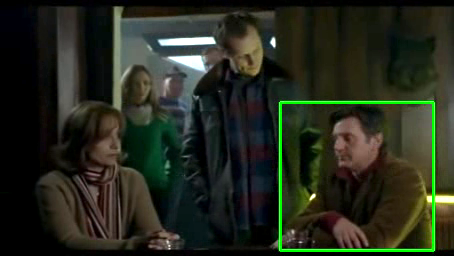}
		&
		\includegraphics[height=\imageheight, trim={2cm 0 0 0}, clip]{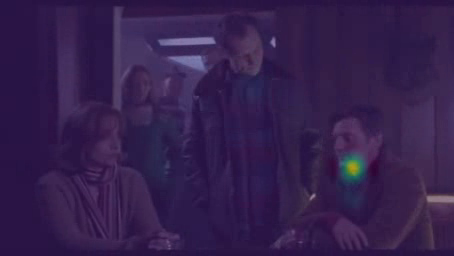}
		&
		\includegraphics[height=\imageheight, trim={2cm 0 0 0}, clip]{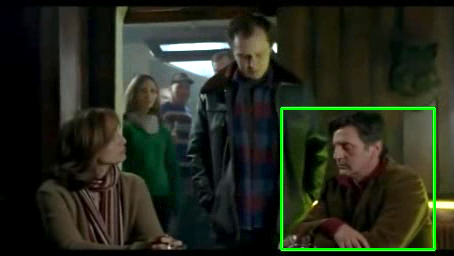}
		&
		\includegraphics[height=\imageheight, trim={2cm 0 0 0}, clip]{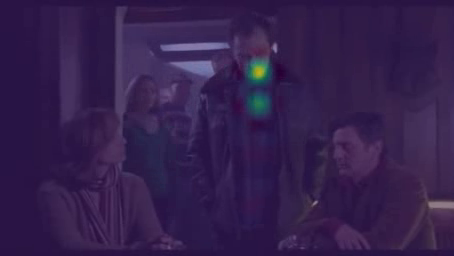}
		&
		\includegraphics[height=\imageheight, trim={0.5cm 0 0.5cm 0}, clip]{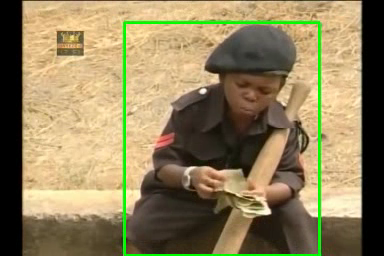}
		&
		\includegraphics[height=\imageheight, trim={0.5cm 0 0.5cm 0}, clip]{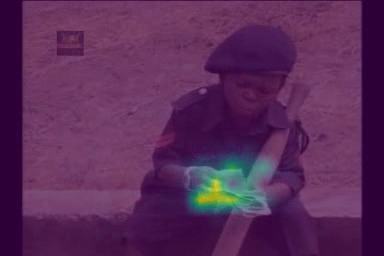}
		&
		\includegraphics[height=\imageheight]{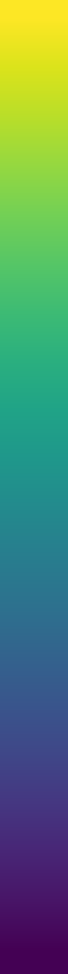}
		\\ 
	\end{tabular}

	\begin{tabularx}{\linewidth}{f{5.6cm}  f{6.5cm} f{4.5cm}}
	\footnotesize{(a)} \scriptsize{\textsf{talk to (0.6), sit (0.92), watch person (0.81)}} & 
	\footnotesize{(b)} \scriptsize{\textsf{listen to person (0.9), watch person (0.89), sit (0.97)}} & 
	\footnotesize{(c)} \scriptsize{\textsf{carry/hold (1.0), sit(0.88)}}
	\end{tabularx}

	\caption{
		Visualisation of spatial message passing when using implicit objects %
		as detailed in Sec.~\ref{sec:exp_qualitative_results}.		
		Note in (a) and (b) that the network focuses on the selected actor (denoted by the green box on the left) when the action is ``talk to'', and on the person speaking to him when the action is ``listen to person''.
		Attention weights, $\alpha_{ij}$ are colour-coded according to the bar on the far right.
		Best viewed in colour.
	}
	\label{fig:implicit_object_gat_vis}
	
	\renewcommand{\arraystretch}{1}  %
	\setlength{\tabcolsep}{6pt}  %
	\vspace{-0.5\baselineskip}
\end{figure*}

We observe from the second row of Tab.~\ref{tab:ablation_ava_message_passing} that passing messages only between actor nodes (and thus not modelling object interactions) provides an overall improvement of 0.8 points.
The largest gain, as expected, are for Human-Human action classes, as these are the interactions modelled by passing messages between actor nodes in the graph.
When passing messages from implicit context nodes to the actors (third row), we observe an improvement in all types of action classes, but primarily in Human-Human and Human-Object classes.
This is because implicit context nodes encompass the entire feature map, and thus capture information about the whole scene (as also shown in Fig.~\ref{fig:implicit_object_gat_vis}).

Modelling objects explicitly with external regional proposals (fourth row) improves on the same action types as the implicit object model, but performs marginally worse.
This suggests that our RPN trained on OpenImages~\cite{kuznetsova_ijcv_2020} is unable to detect the objects  that are most discriminative of AVA actions.
As object annotations are not provided in AVA, it is not possible to evaluate the recall of our external region proposals on relevant objects.
It is not clear if we should expect more improvement from explicit objects, because although previous works have considered implicit~\cite{girdhar_cvpr_2019,sun_eccv_2018,zhang_tokmakov_cvpr_2019} and explicit representations~\cite{baradel_eccv_2018, wang_eccv_2018}, we are not aware of any that have compared the two.
Note that these explicit context nodes also model Human-Human interactions as our RPN is trained on people in OpenImages~\cite{kuznetsova_ijcv_2020}.  %
We obtain further improvements by combining messages from actors, implicit and explicit context nodes, as shown by the next two rows. %

Finally, we evaluate our spatio-temporal model, which passes messages from actors, implicit- and explicit-context nodes with $\tau_c = 3$ and $\tau_s = 3$, corresponding to an overall temporal window of 8.5 seconds.
This model performs the best, improving upon the baseline by 2.2 points, or a relative improvement of 8.9\%.
In particular, the temporal connections help to improve on Human-Object action classes.

AVA is a long-tailed dataset, and the most common failure modes of all variants of our model are the classes with few training examples.
However, our model improves on both head and tail classes with respect to the baseline.
This, and detailed per-class results, are in the appendix.

\begin{table}[tb]
	\centering
	\caption{Comparison on AVA. 
		We report the Mean AP using v2.1 and v2.2 annotations.
		All methods pretrained on Kinetics 400.
		``Multiscale'' refers to averaging results over three scales~\cite{wu_cvpr_2019}.
	}
	\small{
		\begin{tabular}{lcc}
			\toprule
			Method					 								& v2.1 & v2.2 \\ \midrule
			ACRN (S3D) \cite{sun_eccv_2018} 						& 17.4 & --\\
			Zhang \etal (R50) \cite{zhang_tokmakov_cvpr_2019} & 22.2 & -- \\
			SlowFast baseline (R50) 		& 24.5 & 24.8 \\ %
			Girdhar~\etal (I3D) \cite{girdhar_cvpr_2019}		& 25.0 & -- \\
			LFB (R50) \cite{wu_cvpr_2019}						& 25.8 & -- \\
			Ours (R50)										& \textbf{26.5} & \textbf{27.0} \\ %
			\midrule
			Ours Multiscale (R50)							&	\textbf{27.3} & \textbf{27.7} \\
			\midrule
			SlowFast baseline (R101)	& 26.3 & 26.7\\ %
			LFB (R101) \cite{wu_cvpr_2019}					& 26.8 & -- \\
			Ours (R101) & \textbf{28.3} & \textbf{28.8} \\ %
			\midrule
			LFB Multiscale (R101)~\cite{wu_cvpr_2019}			& 27.7 & -- \\
			Ours Multiscale (R101)							& \textbf{29.5} & \textbf{30.0} \\
			\bottomrule
		\end{tabular}
	\vspace{-\baselineskip}
		\label{tab:ava_sota_comparison}
	}
\end{table}

\vspace{\paravspace}
\paragraph{State-of-the-art-comparison on AVA}
Table~\ref{tab:ava_sota_comparison} compares to recent, published work on AVA.
Our method builds upon SlowFast~\cite{feichtenhofer_iccv_2019} as the base architecture, and our graph model shows substantial improvements with either a 3D ResNet 50 or ResNet 101 backbone.
As discussed in Sec.~\ref{sec:method_discussion}, 
 Girdhar~\etal~\cite{girdhar_cvpr_2019} and LFB~\cite{wu_cvpr_2019} can be considered as special cases of our graph model as they model only spatial and temporal edges respectively.
Our method, which constructs a spatio-temporal graph, outperforms both.
Note that we outperform LFB~\cite{wu_cvpr_2019} across both ResNet 50 and ResNet 101 backbones.
Zhang~\etal~\cite{zhang_tokmakov_cvpr_2019} also model a spatio-temporal interactions, but employ three separate graphs with a hand-crafted aggregation function which we also outperform with our single, coherent spatio-temporal graph.

\begin{figure*}[tb]
	\centering
	\renewcommand{\arraystretch}{0.8}  %
	\setlength{\tabcolsep}{2pt} %
	
	\begin{tabular}{cl}
		\includegraphics[width=0.98\linewidth]{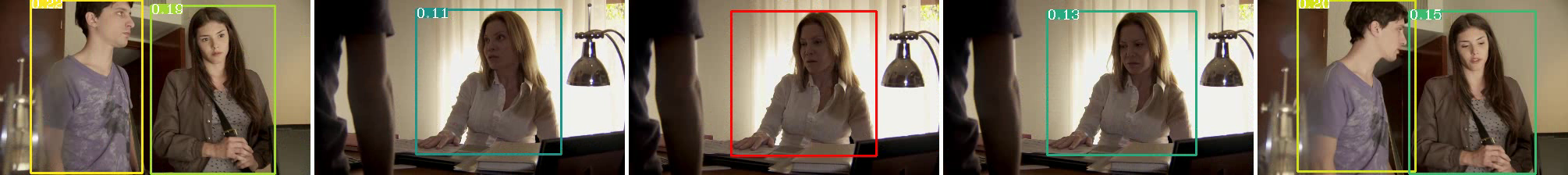}
		&
		\includegraphics[height=1.90cm]{figures/gat_visualisations/viridis}
		\\ 
		\scriptsize{\textsf{talk to (0.85), watch a person (0.82), sit(0.51)}} \\
	\end{tabular}
	\caption{Visualisation of temporal graph as detailed in  Sec.~\ref{sec:exp_qualitative_results}.
		We show the neighbours for the actor depicted by the red box in the centre keyframe, colour-coding the respective boxes with their attention weights ($\alpha_{ij}$).
		Predicted action scores of the red actor are below, and selected keyframes are $\tau_s = 2.1$ seconds apart.
		Notice that the model focuses on the people 2 keyframes (4.2 seconds) on either side of the actor to recognise that she is ``talking to'' and ``watching'' them.
	}
	\label{fig:temporal_gat_vis}
	
	\renewcommand{\arraystretch}{1}  %
	\setlength{\tabcolsep}{6pt}  %
	\vspace{-1\baselineskip}
	
\end{figure*}

\vspace{\paravspace}
\paragraph{State-of-the-art comparison on UCF101-24}
Table~\ref{tab:ucf_sota_comparison} shows that we outperform recent, published work on UCF101-24~\cite{soomro_arxiv_2012}, using either a 3D ResNet 50 or ResNet 101 backbone.
We do so without using optical flow as an additional input modality, showing that our network can capture temporal information without it.
Moreover, the improvements obtained from our graph model are consistent with our results on the Action Genome and AVA datasets.

\begin{table}[tb]
	\caption{Comparison to state-of-the-art on UCF101-24. We report the Frame AP at 0.5 using the corrected annotations of~\cite{singh_iccv_2017}.}
	\centering
	\small{
		\begin{tabular}{lcc}
			\toprule
			Method		   										& Modality 			& Mean AP \\ \midrule
			ACT~\cite{kalogeiton_iccv_2017} 	  & RGB + Flow     &   69.5       \\
			Song~\etal~\cite{song_cvpr_2019}	&RGB + Flow		 &   72.1	    \\
			STEP~\cite{yang_cvpr_2019}				& RGB + Flow	 &   75.0		\\
			Gu~\etal~\cite{gu_cvpr_2018}		   & RGB + Flow		&   76.3	   \\
			MOC~\cite{li_moc_eccv_2020} 		  & RGB + Flow	  &   78.0		 \\
			\midrule
			SlowFast (R50)						& RGB			   &  76.6 \\
			SlowFast (R101)					   & RGB			  &  77.4 \\
			Ours (R50)							  & RGB				 &  78.6 \\
			Ours (R101)							 & RGB				&  \textbf{79.3} \\
			\bottomrule
		\end{tabular}
	}
	\label{tab:ucf_sota_comparison}
	\vspace{-\baselineskip}
\end{table}

\paragraph{Qualitative results}
\label{sec:exp_qualitative_results}

We visualise the spatial and temporal messages received by an actor node in the graph, on the AVA dataset, when using GAT as the message passing function.
In particular, we visualise the attention weights, $\alpha_{ij}$ in \eqref{eq:att_weights}, on each implicit object node (Fig.~\ref{fig:implicit_object_gat_vis}) and actor in neighbouring keyframes (Fig.~\ref{fig:temporal_gat_vis}), for a given actor.
We observe that the network places more weight on graph nodes that are intuitively consistent with the final action prediction.
For example, in Fig.~\ref{fig:implicit_object_gat_vis}, the model focuses on the actor's face when his action is ``talk to'' and the other person's face when his action is ``listen to''.

\section{Conclusion and Future Work}

We have presented a novel spatio-temporal graph neural network framework to explicitly model interactions between actors, objects and their environment.
Our formulation can model objects either implicitly or explicitly, 
and generalises previous structured models for video understanding~\cite{girdhar_cvpr_2019, sun_eccv_2018, wu_cvpr_2019, zhang_tokmakov_cvpr_2019}.
Using our versatile approach, we have achieved state-of-the-art results
on two diverse tasks across three datasets.
Future work remains to harness explicit object representations more effectively on AVA.

{\small
\bibliographystyle{ieee_fullname}
\bibliography{bibliography}
}

\clearpage
\appendix

\section*{Appendix}

In this appendix, we additional analysis on AVA (Sec.~\ref{sec:supp_ava_per_class}) and implementation details (Sec.~\ref{sec:supp_implementation_details}).
\section{AVA per-class analysis}
\label{sec:supp_ava_per_class}

\begin{figure*}
	\includegraphics[width=1.02\linewidth]{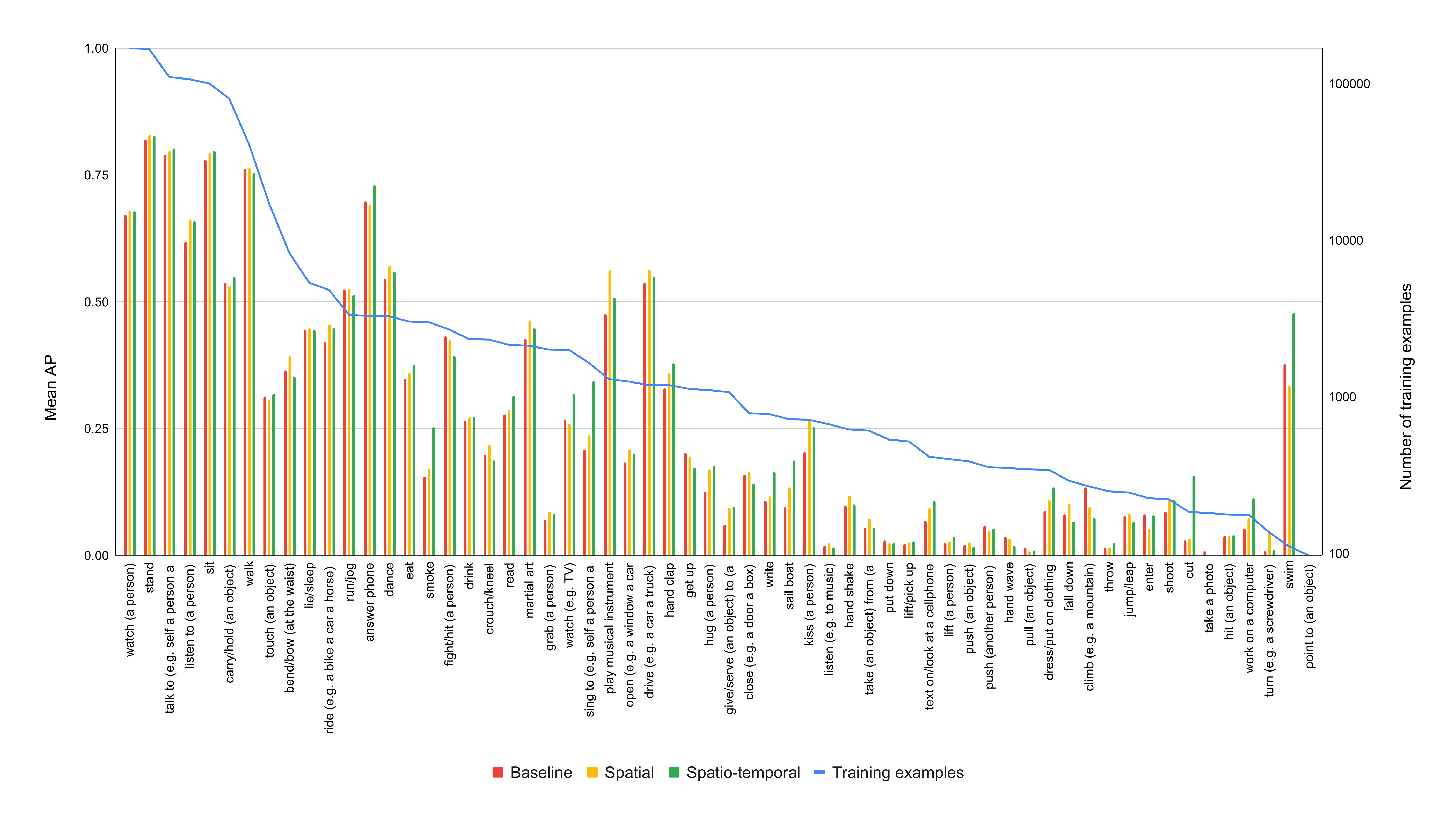}
	\caption{Mean Average Precision for each action class on AVA, for the SlowFast baseline, spatial- and spatio-temporal graph models with a ResNet-50 backbone.
	The number of examples in the training set are shown with the blue line on the right, vertical axis. Note that a logarithmic scale is used, as AVA has a long-tailed distribution of labels.
	}
	\label{fig:ava_per_class_all}
\end{figure*}

Figure~\ref{fig:ava_per_class_all} presents per-class results on AVA for our SlowFast baseline, spatial and spatio-temporal graph models when using a ResNet-50 backbone.
We additionally show the number of examples in the training set.
Our proposed graph models improve on both head and tail classes in AVA.
In fact, the largest absolute improvements are observed on tail classes such as ``cut'', ``swim'' and ``sing to''.
However, note that the absolute accuracy of each class is still correlated with the number of training examples.
The classes which all variants of our model perform the worst on, are the classes with the least training examples such as ``point to (an object)'', ``take a photo'' and ``turn (eg a screwdriver)''.
To emphasise the long-tailed distribution of AVA, note that ``point to (an object)'' has the fewest training examples (just 97), whilst ``watch (a person)'' has the most with 168 148.
\section{Additional Implementation details}
\label{sec:supp_implementation_details}

\paragraph{Person detections}

For our spatio-temporal action detection experiments, we use external person detections as our actor region proposals like~\cite{feichtenhofer_iccv_2019, wu_cvpr_2019}.
For our experiments on AVA, we use the person detections publicly released by~\cite{feichtenhofer_iccv_2019, wu_cvpr_2019} during both training and testing.
More specifically, this is a Faster-RCNN~\cite{ren_neurips_2015} model, pretrained on Microsoft COCO~\cite{lin_coco_eccv_2014} and then finetuned on the training set of AVA~\cite{gu_cvpr_2018}, using Detectron~\cite{detectron}.

For our experiments on UCF101-24~\cite{soomro_arxiv_2012}, we finetune a Faster-RCNN detector pretrained on COCO on the UCF101-24 training set using Detectron.

When training on both AVA and UCF101-24 datasets, we use both ground-truth and predicted person bounding boxes.
Predicted bounding boxes are assigned labels by matching them to ground-truth boxes using an IoU threshold of 0.75.
Predicted bounding boxes which do not match any ground truth boxes act as negative examples for all action classes.

\paragraph{OpenImages object detector}
For our ``explicit object representation'' experiments on AVA, we use the publicly available Faster-RCNN detector trained on OpenImages v4\footnote{\url{https://tfhub.dev/google/faster_rcnn/openimages_v4/inception_resnet_v2/1}}.

\paragraph{Optimiser hyperparameters}
We train our models following the settings of the publicly released SlowFast code~\cite{feichtenhofer_iccv_2019}.
We train for 20 epochs using synchronous Stochastic Gradient Descent (SGD) and a momentum of 0.9 on 8 Nvidia V100 (16 GB) GPUs.
The learning rate was set to 0.1, and reduced by a factor of 10 after 10 and 15 epochs respectively.
We employed a linear-warmup schedule for the learning rate, increasing linearly from $1.25 \times 10^{-4}$ to $0.1$ in the first 5 epochs.
We also used a weight decay of $10^{-7}$.
An epoch is defined as all the keyframes in the dataset.

\paragraph{Training loss functions}
For spatio-temporal action recognition on AVA and UCF101-24, we use the binary cross-entropy loss function.
The loss function for a single example is:
\begin{equation}
L(x, y) = -\frac{1}{C}\sum_{i}^{C} y_i \log{\left(\sigma(x_i)\right)} + (1 - y_i)\log{\left(1 - \sigma(x_i)\right)},
\end{equation}
where $C$ is the number of classes, $\sigma$ is the sigmoid activation function, $x \in \mathbb{R}^C$ denotes the logits predicted by the network and $y_i \in \{0, 1\}$ denotes the binary ground truth for the $i^{th}$ class.

For scene graph classification, we use two loss functions: One for predicting the object class of each bounding box proposal, and another for predicting the relationship label between each of the $\frac{N(N-1)}{2}$ pairs of object proposals.
For the former, we use the softmax cross-entropy, as each proposal can only be assigned a single object class.
And for the latter, we use the sigmoid cross-entropy, as there can be multiple relationship labels between any pair of objects.
Concretely, the loss is
\begin{align}
L(x, w, y, z) &= \lambda L_{\text{object}}(x, y) + L_{\text{rel}}(w, z) \\
L_{\text{object}}(x, y) & = -\frac{1}{N}\sum_{i}^{N}\sum_{j}^{C}  y_{ij} \log\left( \text{Softmax}(x)_{ij} \right) \\
L_{\text{rel}}(w, z) &= 	- \frac{2}{N(N-1)R} \sum_{i}^{N} \sum_{j}^{i - 1} \sum_{r}^{R}  z_{ijr} \log\left( \sigma(w_{ijr}) \right) \nonumber \\
& + (1 - z_{ijr})\log\left( 1 - \sigma(w_{ijr}) \right).
\end{align}
Here, $N$ is the number of object proposals in the video clip,  $R$ the number of relationship classes and $C$ the number of object classes.
$y$ is the one-hot, ground-truth object class label, $z$ the binary ground-truth relationship label, and $x$ and $w$ denote the object- and relationship-logits respectively.
On Action Genome~\cite{ji_cvpr_2020}, there are $R =25$ relationship classes, and $C = 35$ object classes.
We set $\lambda = 0.5$ during training on Action Genome to prevent the object class loss from dominating the overall loss (the softmax cross entropy has a higher loss than the binary sigmoid cross entropy at the start of training).

\begin{table}[tb]
    \centering
    \caption{Ablation study of the number of heads when using GAT as the message passing function.
    Mean AP reported on AVA using a ResNet-50 backbone.
    }
    \begin{tabular}{lccccc}
    \toprule
    Number of heads & 1 & 2 & 3 & 4 & 5 \\
    \midrule
    Mean AP & 25.4 & 25.8 & 25.9 & 25.9 & 25.9 \\
    \bottomrule
    \end{tabular}
    \label{tab:ablation_gat_heads}
\end{table}

\paragraph{Multi-headed attention}
For the GAT~\cite{velivckovic_iclr_2018} and Non-Local~\cite{wang_cvpr_2018} message-passing functions used in the paper (Sec 3.4), it is common to compute ``multi-headed'' attention~\cite{vaswani_neurips_2017}.
In all experiments reported in the main paper, we used 4 heads, as motivated by the ablation study using GAT on the AVA dataset in Tab.~\ref{tab:ablation_gat_heads}.
To combine the messages from each head, we performed an attention-weighted convex combination as done in Eq.~\ref{eq:att_weights}.

\paragraph{Multiscale testing}
We use three scales when performing multiscale testing (as done in Table 3 of the main paper).
Specifically, we resize input video-clips so that the shortest side is 224, 256 and 320 respectively.

\end{document}